\documentclass[conference]{IEEEtran}
\IEEEoverridecommandlockouts

\usepackage{cite}
\usepackage{amsmath,amssymb,amsfonts}
\usepackage{algorithmic}
\usepackage{graphicx}
\usepackage{textcomp}
\usepackage{xcolor}
\usepackage{hyperref} 
\usepackage{algorithm}
\usepackage{algorithmic}

\begin{document}

\title{IsoSignVid2Aud: Sign Language Video to Audio Conversion without Text Intermediaries}

\author{\IEEEauthorblockN{Harsh Kavediya$^{\dagger,1}$, Vighnesh Nayak$^{\ddagger,1}$, Bheeshm Sharma$^\star$ and Balamurugan Palaniappan$^\mathsection$}
	\IEEEauthorblockA{\textit{$^{\dagger,\ddagger}$Department of Mechanical Engineering, $^{\star,\mathsection}$ Department of IEOR} \\
		\textit{Indian Institute of Technology Bombay}, Mumbai, India. \\ 
		email: $^\dagger$\texttt{harshkavediya2003@gmail.com}, $^\ddagger$\texttt{vighneshnayak203@gmail.com}, \\ $^\star$\texttt{bheeshmsharma@iitb.ac.in}, $^\mathsection$ \texttt{balamurugan.palaniappan@iitb.ac.in}} \\
        \thanks{$^1$Denotes equal contribution}
    }

	

\maketitle

\begin{abstract}
Sign language to spoken language audio translation is important to connect the hearing- and speech-challenged humans with others. We consider sign language videos with isolated sign sequences rather than continuous grammatical signing. Such videos are useful in educational applications and sign prompt interfaces. Towards this, we propose 
IsoSignVid2Aud, a novel end-to-end framework that translates sign language videos with a sequence of possibly non-grammatic continuous signs to speech without requiring intermediate text representation, providing immediate communication benefits while avoiding the latency and cascading errors inherent in multi-stage translation systems.  Our approach combines an I3D-based feature extraction module with a specialized feature transformation network and an audio generation pipeline, utilizing a novel Non-Maximal Suppression (NMS) algorithm for the temporal detection of signs in non-grammatic continuous sequences. Experimental results demonstrate competitive performance on ASL-Citizen-1500 and WLASL-100 datasets with Top-1 accuracies of 72.01\% and 78.67\%, respectively, and audio quality metrics (PESQ: 2.67, STOI: 0.73) indicating intelligible speech output. Code is available at: \href{https://github.com/BheeshmSharma/IsoSignVid2Aud_AIMLsystems-2025}{this link}.
\end{abstract}

\section{Introduction}
Sign languages are the primary means of communication for millions of deaf and hard-of-hearing individuals worldwide. Despite this, significant communication barriers persist between signing and non-signing communities, often requiring human interpreters or multi-stage technological solutions that introduce latency and errors. Traditional sign language processing systems typically follow a two-stage approach: first recognizing signs using computer vision techniques, then translating the recognized signs into text before optional conversion to speech \cite{carreira2017quo,bohacek2022sign}. This multi-stage pipeline introduces compounding errors and often fails to capture the rich expressiveness inherent in sign languages, limiting its generalizability, especially for complex signs.

Recent advances in deep learning have enabled significant progress in both sign language recognition and speech synthesis independently. Models like I3D \cite{carreira2017quo} have demonstrated impressive capabilities in capturing the spatio-temporal dynamics crucial for sign language understanding, while modern speech synthesis techniques have achieved increasingly natural-sounding output. However, these advancements have largely developed in parallel, with limited exploration of direct sign-to-speech conversion.



Our approach is motivated by the natural process of language acquisition, drawing a deliberate parallel, grounded in developmental linguistics. Just as children first master individual words like \textit{mama}, \textit{dada}, or \textit{milk} before combining them into sentences, sign language learners follow a similar developmental path: mastering isolated signs before progressing to complex grammatical structures. By focusing on isolated signs, we target a critical stage in communication development where meaning can already be conveyed effectively even without full grammatical fluency. This choice is not a limitation, but a strategic one: isolated signs allow early learners and non-signers to engage in meaningful, real-time interactions from the very beginning. While continuous signing captures the full richness of sign language, processing isolated signs addresses a complementary and equally important challenge, establishing foundational communication bridges that are both accessible and immediately useful.


In this paper, we present IsoSignVid2Aud, a novel end-to-end framework that directly converts sign language videos to speech without requiring an intermediate text representation. Our system is specifically designed to recognize sequences of isolated signs, making it well-suited for high-impact practical applications where isolated recognition delivers immediate value: emergency communication and keyword recognition in public spaces (identifying signs like ``help", ``bathroom" or ``information"), accessibility interfaces for diverse users, command-based systems for menu navigation in kiosks and smart devices, and other scenarios that prioritize practical, real-world utility over linguistic completeness.

The IsoSignVid2Aud system consists of three primary components: a feature extraction module based on the I3D network to capture spatio-temporal information from sign language videos, a feature transformer that maps visual features to audio representations in the form of spectrograms, and an audio generation module that synthesizes natural-sounding speech from the predicted spectrograms.

We evaluate our proposed framework on two standard datasets: ASL-Citizen-1500 and WLASL-100. Our experiments demonstrate that IsoSignVid2Aud achieves competitive performance in both sign recognition accuracy and speech quality. These results suggest that direct sign-to-speech conversion based on isolated sign recognition is not only feasible but potentially advantageous compared to traditional multi-stage approaches that require continuous grammatical signing.

To the best of our knowledge, we are the first to propose a direct sign language video-to-audio model without the need for intermediate text. This contributes a new paradigm for sign-to-speech translation that bridges the gap between early-stage sign language understanding and accessible communication technologies, particularly for non-signers and emerging sign learners. Another key innovation in our approach is the introduction of a Non-Maximal Suppression (NMS) algorithm for temporal sign isolation, enabling our system to process sign language videos effectively by identifying and isolating the most significant sign gestures while eliminating redundant or low-confidence predictions, which aligns with how humans naturally acquire and process sign language by recognizing distinct signs before interpreting their combined meaning. 


\section{Related Work}

\subsection{Sign Language Recognition}
Sign language recognition has seen significant advancements with the development of deep learning techniques. Early approaches relied on handcrafted features \cite{koller2016deep}, but recent methods leverage deep neural networks for more robust feature extraction. 

Convolutional neural networks (CNNs) have been widely adopted for this task, with notable architectures including I3D \cite{carreira2017quo}, which has demonstrated strong performance in action recognition tasks and sign language recognition. I3D's ability to capture spatio-temporal information makes it particularly suitable for understanding the dynamic nature of sign language gestures.

More recently, transformer-based approaches have shown promising results. SignBERT \cite{hu2021signbert} introduced a pre-training approach for hand-model-aware representation. SiFormer \cite{jiang2023siformer} proposed a spatiotemporal transformer specifically designed for sign language recognition, achieving state-of-the-art results on benchmark datasets. Spoter \cite{bohacek2022sign} presented a sign pose-based transformer for word-level recognition, while UniSign \cite{hu2023unisign} introduced a universal sign language recognition model that achieves superior performance across datasets.

Graph-based methods have also been explored, with ST-GCN \cite{yan2018spatial} applying graph convolutional networks to skeleton-based action recognition, which has been adapted for sign language processing. StepNet \cite{Shen_2024} introduced a spatial-temporal part-aware network specifically designed for isolated sign language recognition.

\subsection{Sign Language Translation}
Traditional sign language translation systems typically follow a two-stage approach: first recognizing signs using computer vision techniques, then translating the recognized signs into text before optional conversion to speech \cite{camgoz2018neural}. This multi-stage pipeline often introduces compounding errors and fails to capture the expressiveness of sign languages.

Recent research has focused on end-to-end approaches for sign language translation. Camgoz et al. \cite{camgoz2020sign} proposed an end-to-end sign language translation network using transformer architectures. Khandelwal et al. \cite{khandelwal2022simpleeffectiveclipembeddings} introduced a simple yet effective approach using temporal convolutions for sign language translation. Ko et al. \cite{ko2019neural} explored neural sign language translation using sequence-to-sequence models with attention mechanisms.

\subsection{Speech Synthesis}
Speech synthesis has evolved significantly with deep learning approaches. Traditional concatenative and parametric methods have been largely replaced by neural vocoders such as WaveNet \cite{oord2016wavenet}, which generate high-quality speech directly from text or linguistic features. 

More efficient approaches like HiFi-GAN \cite{kong2020hifi} utilize generative adversarial networks to produce high-fidelity speech with fewer computational requirements. This model employs multi-period and multi-scale discriminators to capture both local and global structures in the audio waveform, producing natural-sounding speech with fewer artifacts compared to traditional methods.

Tacotron 2 \cite{shen2018natural} combines a sequence-to-sequence model with a WaveNet vocoder to generate natural speech from text, while FastSpeech \cite{ren2019fastspeech} introduced a non-autoregressive model for fast, high-quality speech synthesis.

\subsection{Temporal Detection in Continuous Signing}
Detecting and segmenting signs in continuous videos remains challenging. Pu et al. \cite{pu2019iterative} proposed an iterative alignment network for continuous sign language recognition. The use of Non-Maximal Suppression (NMS) for temporal detection, as implemented in our work, builds upon techniques from object detection \cite{girshick2015fast} and adapts them specifically for sign localization in the temporal dimension.

Continuous sign language recognition approaches often employ recurrent neural networks \cite{cui2019deep} or attention mechanisms \cite{zhou2020spatial} to model the temporal dependencies between signs. Recent work by Min et al. \cite{min2021visual} explored visual alignment constraints for continuous sign language recognition.

Our IsoSignVid2Aud approach differs from previous work by proposing a direct end-to-end framework for converting sign language videos to speech without requiring intermediate text representation, combining advanced feature extraction techniques with specialized audio generation pipelines to preserve more of the expressiveness inherent in sign languages.

\section{Methodology}

\begin{figure*}
    \centering
    \includegraphics[scale=0.14]{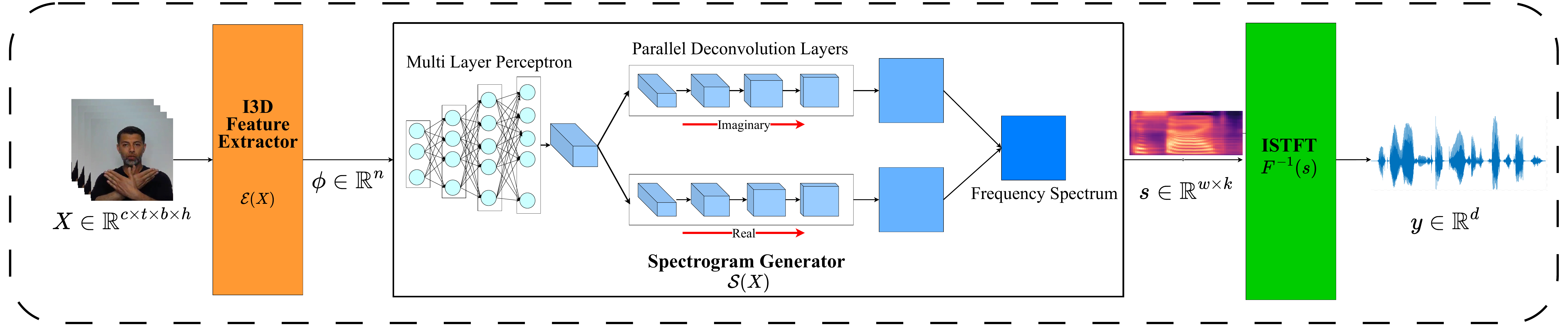}
    \caption{Proposed Architecture of IsoSignVid2Aud \\ (Video frames are a part of ASL-Citizen dataset \cite{desai2023asl})}
    \label{fig:system}
\end{figure*}

Our IsoSignVid2Aud framework 
consists of three major components: (1) a feature extraction module based on I3D network, (2) a feature transformer that maps visual features to audio representations, and (3) an audio generation module to synthesize natural-sounding speech. Figure~\ref{fig:system} presents an overview of our end-to-end pipeline.

\subsection{System Architecture}

The IsoSignVid2Aud framework operates modularly to achieve direct conversion from sign language videos to speech. We first extract deep spatio-temporal features from sign language videos using an I3D \cite{carreira2017quo} model pretrained on the Kinetics dataset \cite{kay2017kineticshumanactionvideo} and fine-tuned on the ASL-Citizen \cite{desai2023asl} and WLASL \cite{li2020word} datasets. These features then transform our Spectrogram Generator, comprising a multi-layer perceptron and transposed convolution network, to produce the magnitude and phase of mel spectrograms. Then these synthesized spectrograms are converted to audible speech using the Inverse Short-Time Fourier Transform (ISTFT) algorithm \cite{griffin1984signal}.

\subsection{Feature Extraction}


To extract meaningful features from sign language videos, we require a model capable of capturing both spatial and temporal dynamics to accurately interpret the expressive and sequential nature of sign language. To meet these needs, we employ the Inception 3D (I3D) model \cite{carreira2017quo} pre-trained on the Kinetics dataset \cite{kay2017kineticshumanactionvideo} and fine-tuned on the ASL Citizen and WLASL datasets for extracting meaningful features from sign language videos. The I3D architecture, an extension of the Inception model \cite{szegedy2014goingdeeperconvolutions}, with all 2D convolutions inflated to 3D convolutions by repeating filter kernel parameters along the temporal dimension, enables direct processing of video data. This approach allows I3D to utilize pre-trained 2D image classification weights while effectively modeling spatio-temporal data.

The I3D model processes an input video $X \in \mathbb{R} ^{c\times t\times b\times h}$, where $c$ denotes the number of channels in the input video, $t$ is the duration of the video, and $b$, $h$ are the frame width and height, respectively. The I3D model outputs a feature vector $\phi \in \mathbb{R}^{n}$ along with a confidence vector $\mathcal{C} \in \mathbb{R}^k$ which can be expressed as
\begin{equation} 
 \phi, \mathcal{C} = \mathcal{E}(X) 
\end{equation}
where $n$ denotes the dimensionality of the feature representation of the video, and $k$ is the number of classes. 

\subsection{Spectrogram Generation}

The Spectrogram Generator module is a key component of our approach. It bridges visual features and audio representations, enabling our model to function independently of text by converting visual inputs into spectrogram-based audio features. Due to the absence of a pre-existing feature-to-spectrogram matching dataset, we train the spectrogram generation model on a custom dataset. This dataset is generated by creating spectrograms from the Tacotron2 \cite{oord2016wavenet} model using the text glosses from the WLASL and ASL-Citizen datasets, which serve as the ground truth.

During training, we utilize the spectrogram magnitudes and phases from the custom dataset and map them to the corresponding I3D features, following the training of the I3D feature extractor. This spectrogram generator component processes the feature vector $\phi\in\mathbb{R}^{n}$ obtained from the I3D model $\mathcal{E}$  and utilizes a multi-layer perceptron followed by a transposed convolution network $\mathcal{S}$ to map it to the real and imaginary parts of the spectrogram,  each of the dimensions $r,i\in\mathbb{R}^{\omega\times \tau}$ as in Figure \ref{fig:system}, where $\omega$ and $\tau$ denote the number of frequencies and time steps respectively. The functionality of the Spectrogram Generator can be expressed as 
\begin{equation}
    r,i = \mathcal{S}(\phi).
\end{equation} 
The multilayer perceptron reduces the feature dimensions, which are then reshaped and upsampled by the transposed convolution layers to obtain relatively larger spectrograms. The architectures of different components are given in Section 4.

\subsection{Audio Generation}
Our pipeline's final stage converts predicted spectrograms into audible speech via the Inverse Short-Time Fourier Transform (ISTFT). Utilizing the spectrogram's real (\( r \)) and imaginary (\( i \)) components, ISTFT directly reconstructs the time-domain signal. Unlike vocoder-based approaches requiring additional neural models, ISTFT provides an efficient, high-fidelity solution for audio generation from complex spectrograms.

Given a spectrogram with real and imaginary components \( r \in \mathbb{R}^{\omega \times \tau} \) and \( i \in \mathbb{R}^{\omega \times \tau} \), respectively, we generate audio \( y \in \mathbb{R}^d \) using the ISTFT applied to the complex spectrogram \( s \) constructed from \( r \) and \( i \). The ISTFT operation is expressed as:
\begin{equation}
y = \mathcal{F}^{-1}(s),
\end{equation}
where \( \mathcal{F}^{-1} \) denotes the inverse Fourier transform.

\subsection{Non-Maximal Suppression for Temporal Detection}

We propose a Non-Maximal Suppression (NMS) algorithm for temporal sign isolation. This is particularly important as our framework is designed to identify single signs, while real-world sign language videos often contain multiple signs in sequence. The NMS module ranks predictions based on confidence and imposes temporal constraints such that the final audio output is consistent.

Algorithm~\ref{alg:nms} outlines our approach for selecting the most reliable feature representations from streaming video frames.

\begin{algorithm}[t]
\caption{Temporal Non-Maximal Suppression}
\label{alg:nms}
\begin{algorithmic}[1]
\REQUIRE Frame stream, window size $w$, hop length $h_\ell$, overlap threshold $o$, confidence threshold $\theta$
\ENSURE Feature vectors of detected signs

\STATE \textbf{Notations:} 
\STATE \hspace{\algorithmicindent} $W$: Window (deque with max size $w$)
\STATE \hspace{\algorithmicindent} $\mathcal{E}(W)$: Feature extraction function returning $(\phi, \mathcal{C})$
\STATE \hspace{\algorithmicindent} $\phi$: Feature vector
\STATE \hspace{\algorithmicindent} $\mathcal{C}$: Confidence score vector

\STATE $W \leftarrow \emptyset$, $t_{best} \leftarrow \text{null}$, $\phi_{best} \leftarrow \text{null}$, $\mathcal{C}_{best} \leftarrow 0$, $t \leftarrow 0$
\STATE \textbf{Note:} $W$,  $t_{best}$,  $\phi_{best}$, $C_{best}$ are global variables. 
\FORALL{new frames $f$ in stream}
    \STATE Add $f$ to $W$
    \IF{$|W| < w$}
        \STATE \textbf{continue}
    \ENDIF
    
    \STATE $t \leftarrow t + 1$
    
    \IF{$t = 1$}
        \STATE $(\phi, \mathcal{C}) \leftarrow \mathcal{E}(W)$
        \IF{$\max(\mathcal{C}) \geq \theta$}
            \STATE $t_{best} \leftarrow t$, $\phi_{best} \leftarrow \phi$, $\mathcal{C}_{best} \leftarrow \mathcal{C}$
        \ENDIF
        \STATE \textbf{continue}
    \ENDIF
    
    \IF{$(t-1) \bmod h_\ell \neq 0$}
        \STATE \textbf{continue}
    \ENDIF
    
    \STATE $(\phi, \mathcal{C}) \leftarrow \mathcal{E}(W)$
    
    \IF{$t_{best} = \text{null} \wedge \max(\mathcal{C}) > \theta$}
        \STATE $t_{best} \leftarrow t$, $\phi_{best} \leftarrow \phi$, $\mathcal{C}_{best} \leftarrow \mathcal{C}$
        \STATE \textbf{continue}
    \ENDIF
    
    \IF{$t_{best} = \text{null}$}
        \STATE \textbf{continue}
    \ENDIF
    
    \IF{$t - t_{best} > w - o$}
        \STATE $t_{return} \leftarrow t_{best}$, $\phi_{return} \leftarrow \phi_{best}$
        \IF{$\max(\mathcal{C}) > \theta$}
            \STATE $t_{best} \leftarrow t$, $\phi_{best} \leftarrow \phi$, $\mathcal{C}_{best} \leftarrow \mathcal{C}$
        \ELSE
            \STATE $t_{best} \leftarrow \text{null}$, $\phi_{best} \leftarrow \text{null}$, $\mathcal{C}_{best} \leftarrow 0$
        \ENDIF
        \RETURN $\phi_{return}$
    \ENDIF
    
    \IF{$\max(\mathcal{C}) > \max(\mathcal{C}_{best})$}
        \STATE $t_{best} \leftarrow t$, $\phi_{best} \leftarrow \phi$, $\mathcal{C}_{best} \leftarrow \mathcal{C}$
    \ENDIF
\ENDFOR
\end{algorithmic}
\end{algorithm}

Our NMS implementation processes video frames using a sliding window approach with configurable parameters such as the window size ($w$) which is the number of consecutive frames used for feature extraction, the hop length ($h_\ell$) which is the step size between consecutive windows, the overlap threshold ($o$) which is the minimum separation required between selected windows, and the confidence threshold ($\theta$) which is the minimum confidence score for a prediction to be considered. 

The algorithm maintains a window of consecutive frames and processes new frames as they arrive. For each complete window, our Feature Extractor $\mathcal{E}$ produces a feature vector $\phi$ and confidence score vector $\mathcal{C}$. The algorithm then identifies windows with confidence scores above the threshold and tracks the window with the highest confidence. When a new high-confidence window is sufficiently separated from the current best window (based on the overlap parameter), the best window's features are returned for further processing.

This approach enables our system to identify and isolate the most significant sign gestures in continuous signing while eliminating redundant or low-confidence predictions. By adjusting the window parameters, we can effectively balance detection sensitivity with the natural speed of sign language communication, ensuring that only the relevant sign language features are processed for audio generation.

\subsection{Inference Pipeline}

During inference, we implement a sliding window approach for continuous sign language processing. We process each frame of the video through our `IsoSignVid2Aud` model which orchestrates the entire pipeline:

For each frame of the input video:
\begin{enumerate}
    \item The frame is added to a fixed-size window buffer.
    \item Once the window is full, the I3D module processes the window to produce features and confidence score.
    \item The feature and confidence score are passed to NMS module to identify a valid sign.
    \item The extracted features for the valid sign are passed to the Feature Transformer which converts them to mel-spectrograms.
    \item The ISTFT algorithm then generates the audio output from the predicted spectrograms.
\end{enumerate}

\section{Experimental Setup}

\subsection{Dataset}
We used the ASL-Citizen \cite{desai2023asl} and the WLASL \cite{li2020word} datasets, which contain sign language videos and their corresponding gloss. Each video contains a single signer performing an American Sign Language (ASL) sign, recorded at 25 frames per second. We conducted our experiments on a subset of 1500 of the most frequent glosses of the  ASL-Citizen dataset, which originally consists of around 2731 sign language words. The dataset was split into training, validation, and testing sets of sizes 23365, 5282, and 17798, respectively. We then performed experiments on a subset of the 100 most frequent glosses of the WLASL dataset, which originally consists of around 2000 signs. The dataset was split into training, validation, and testing sets of sizes 1448, 252, and 336, respectively.

\subsection{Data Augmentation}
To improve model robustness and generalization, we performed data augmentation on the videos. For each video, we generated 5 augmented versions with the following transformations.

 \textbf{Spatial augmentations:} For each video we applied a random color jittering with brightness, contrast, and saturation factors of 0.6 respectively, and hue factor of 0.2. This was followed by a random rotation within $[-15^{\circ}, 15^{\circ}]$ range.
  
 \textbf{Temporal augmentations:} We randomly dropped up to $\frac{1}{8}$ of frames to simulate variation in signing speed and encourage temporal robustness.

This increased our effective training set size by a factor of 5, providing greater diversity for model training. All augmented videos were saved as MP4 files using the same encoding as the original dataset.

\subsection{Video Data Processing}
We perform a uniform temporal subsampling of  64 frames from the input video followed by normalization of pixel values to [-1, 1]. We then perform a spatial scaling with short side set to 256 pixels and center crop each frame of the resultant video to $256 \times 256$ pixels. The videos are processed at 25 frames per second.

\subsection{Feature Extractor Architecture}

Our feature extractor utilizes a pre-trained I3D model,
which produces embeddings of dimension 2048 for each video. To get the confidence score, we modify the final block of the model by replacing it with an average pooling operation (kernel size $4 \times 7 \times 7$) to produce spatially condensed feature maps. This is followed by a classification head that consists of a dropout layer ($p=0.5$), a linear projection to the number of classes ($1500$ for ASL Citizen and $100$ for the WLASL dataset), and a ReLU activation. Then we apply an adaptive average pooling and flatten the features of the classifier output. 

\subsection{Feature Extractor Training}

We train the feature extractor using stochastic gradient descent with momentum of 0.9 and $L2$-regularization factor of $10^{-8}$. The initial learning rate is set to $1 \times 10^{-3}$ with a reduce-on-plateau scheduling strategy (factor=0.3, patience=2).

For efficient training, we accumulate gradients over 8 mini-batches before performing parameter updates:

\begin{equation}
\theta_{t+1} = \theta_t - \eta \sum_{i=1}^{8} \nabla_{\theta} \mathcal{L}(x_i, y_i; \theta_t)
\end{equation}
where $\theta$ represents model parameters, $\eta$ is the learning rate, and $\mathcal{L}$ is the cross-entropy loss function.

The model was trained for a maximum of 100 epochs with early stopping (patience=5) monitoring validation loss. We utilized a batch size of 4 and processed the data using 4 worker threads on a dedicated CUDA device.

\subsection{Spectrogram Generator Architecture}
Our spectrogram generator transforms video features extracted by the I3D model into complex spectrograms using a multi-layer perceptron (MLP) followed by dual deconvolution networks for the magnitude and phase components. The model takes input features of dimension $2048$ from the I3D model. The MLP changes dimensions progressively $[648, 1296, 2592, 5184]$ with layer normalization, LeakyReLU activations, and dropout ($0.1$) between each linear projection. The final MLP output is then reshaped to a $128 \times 9 \times 9$ tensor, representing the base shape for subsequent upsampling which is then passed through two parallel deconvolution networks with identical architectures—one generating the magnitude ($r$) component and the other generating the phase ($i$) component of the spectrogram. Each deconvolution network consists of a series of transposed convolution blocks each of which comprises  a transposed convolution layer followed by instance normalization, ReLU activation, and dropout. 
The blocks successively half the channel dimensions from 64 to 8 by applying kernels of various sizes, stride length and padding.
Both networks conclude with a final transposed convolution with kernel size $3$, stride $1$, and padding $1$, reducing the channel count from $8$ to $1$. The outputs from both networks are concatenated along the channel dimension, resulting in a two-channel spectrogram where the first channel represents the magnitude and the second channel represents the phase with dimensions $ 1025 \times 100$ each. 
\subsection{Feature Generator Training}
The feature generator model employs a specialized loss function that handles the complex nature of spectrograms, with separate components for real and imaginary parts.

The complex spectrogram loss function is defined as:
\begin{equation}
\begin{aligned}
\mathcal{L}_{\text{complex}}(S_{\text{true}}, S_{\text{pred}}) &= 2 \cdot (L1_{\text{real}} + L1_{\text{imag}} + L1_{\text{mag}}) \\
&\quad + \lambda_{\text{sc}} \cdot (SC_{\text{real}} + SC_{\text{imag}}) \\
&\quad + \lambda_{\text{mse}} \cdot (MSE_{\text{real}} + MSE_{\text{imag}})
\end{aligned}
\end{equation}
where $L1_{\text{real}}$ and $L1_{\text{imag}}$ are L1 losses on real and imaginary components, $L1_{\text{mag}}$ is the L1 loss on the reconstructed magnitude, $SC_{\text{real}}$ and $SC_{\text{imag}}$ are spectral convergence losses with $\lambda_{\text{sc}} = 0.5$, and $MSE_{\text{real}}$ and $MSE_{\text{imag}}$ are mean squared error losses with $\lambda_{\text{mse}}$  0. The spectral convergence loss is computed as:

\begin{equation}
\mathcal{L}_{\text{sc}}(M_{\text{true}}, M_{\text{pred}}) = \frac{||M_{\text{true}} - M_{\text{pred}}||_F}{||M_{\text{true}}||_F}
\end{equation}
where $||\cdot||_F$ denotes the Frobenius norm.

The model is optimized using the Adam optimizer with an initial learning rate of $1 \times 10^{-2}$ and weight decay of $1 \times 10^{-8}$. We employ a ReduceLROnPlateau scheduler that reduces the learning rate by a factor of 0.1 after 3 epochs without improvement in validation loss, ensuring adaptive optimization.

For efficient training, we use a batch size of 64 with configurable parallelism through the num\_workers parameter. The training process incorporates an early stopping mechanism with a patience of 10 epochs to prevent overfitting, which can be enabled or disabled via configuration. Training progress is monitored through both training and validation losses, with the best model selected based on the lowest validation loss.

\subsection{Combined Training Procedure}
Our current approach focused on sequential training of the feature extractor and feature transformer models. We also performed a combined training which jointly optimizes the feature extractor and  spectrogram generator models using a unified training pipeline. The feature extractor is optimized using Stochastic Gradient Descent (SGD) with momentum 0.9, while the transformer-based spectrogram generator uses Adam optimizer with an initial learning rate of $1 \times 10^{-2}$ and weight decay of $1 \times 10^{-8}$. We use a ReduceLROnPlateau scheduler that reduces the learning rate by a factor of 0.1 after 3 epochs without improvement in validation loss, ensuring adaptive optimization. We implement a combined loss function that incorporates both classification and spectrogram reconstruction objectives:

\begin{equation}
\mathcal{L}_{\text{combined}} = \mathcal{L}_{\text{cls}} + \lambda_{\text{spec}} \cdot \mathcal{L}_{\text{spec}}
\end{equation}

where $\mathcal{L}_{\text{cls}}$ is the cross-entropy loss for classification, $\mathcal{L}_{\text{spec}}$ is the complex spectrogram loss with the same weighing factors as defined previously, and $\lambda_{\text{spec}} = 2$ is a weighting factor.

For efficient training, we accumulate gradients over 8 batches before applying parameter updates. Different learning rate schedules are implemented for each component: a ReduceLROnPlateau scheduler for the feature extractor that reduces the learning rate when validation loss plateaus, and a CosineAnnealingLR scheduler with a cycle length of 50 epochs for the transformer generator. 
The training process employs an early stopping mechanism with a configurable patience parameter to prevent overfitting. Training progress is monitored through both classification accuracy and validation loss, with the best model selected based on validation performance.

\subsection{Implementation Details}
All models were implemented in PyTorch with the PyTorchVideo library for video processing. Training was performed on a single NVIDIA RTX A6000 GPU with 48GB of memory. For the feature extractor, we used a batch size of 4 due to the memory constraints of 3D CNN processing. For the spectrogram generator, we used a larger batch size of 64. Our complete model contains approximately 100 million trainable parameters and achieves an average processing speed of 22 frames per second (fps).

The complete pipeline utilizes the trained models for inference with non-maximum suppression to filter predictions. We employ a window size of 50 frames with a hop length of 3 and a confidence threshold of 0.7 for reliable detection. 

\subsection{Evaluation Metrics}\label{sec:metrics}
    Since none of the state of the art models deal with audio, we evaluated our approach using \textbf{PESQ} (Perceptual Evaluation of Speech Quality) \cite{rix2001pesq} which measures the naturalness of speech on a scale from -0.5 to 4.5, with higher values indicating better quality; \textbf{STOI} (Short-Time Objective Intelligibility) \cite{taal2011stoi} which quantifies word intelligibility on a scale from 0 to 1, where values closer to 1 represent better intelligibility; \textbf{MSE} (Mean Squared Error) between the generated and the target spectrograms and \textbf{MCD} (Mel Cepstral Distortion) \cite{kubichek1993mcd} which assesses spectral differences between generated and reference audio, with lower values signifying better similarity. We also employed \textbf{Top 1} and \textbf{Top 5} Accuracy to measure word prediction accuracy by transcribing the audio, since we aren't dealing with text intermediaries, alongside \textbf{F1 score}, which provides a balanced measure of precision and recall.

\subsection{Experimental Protocol}
For all experiments, we trained the feature extractor on the augmented training set and validated it on the original validation set. We then extracted features for all videos in training, validation, and test sets using the feature extractor model. This was followed by training the spectrogram generator on extracted features from the training set and validating on the validation set. We then evaluated the complete pipeline on the test set, measuring both classification accuracy and spectrogram quality.

\section{Results and Analysis}

In this section, we evaluate the classification and audio generation performance of our proposed IsoSignVid2Aud model on the ASL-Citizen-1500 and WLASL-100 datasets.

\subsection{Sign Language Recognition Performance}
Table~\ref{tab:classification_results_asl} presents the classification performance of our models on the ASL-Citizen-1500 dataset, which follows the standard train/validation/test splits provided by the dataset creators. IsoSignVid2Aud achieves a Top-1 accuracy of 72.01\% and a Top-5 accuracy of 91.42\%, which is a slight improvement upon the baseline I3D model by 0.91\% and 1.29\%, respectively. Our model also provides an F1 score of 0.72, demonstrating its effectiveness in recognizing ASL signs with both precision and recall. When trained with the combined approach, IsoSignVid2Aud achieves even better results with a Top-1 accuracy of 75.19\%, Top-5 accuracy of 92.65\%, and an F1-Score of 0.75. This represents a significant improvement of 4.09\% in Top-1 accuracy over the baseline I3D model, suggesting that the joint training approach enhances the feature extraction capabilities for sign language recognition.

Recent advancements in sign language recognition have yielded a diverse set of architectures with varying approaches to feature extraction and representation. Transformer-based models such as UniSign~\cite{hu2023unisign}, SiFormer~\cite{jiang2023siformer}, and SignBERT~\cite{hu2021signbert} leverage self-attention mechanisms to capture complex temporal dependencies in sign language sequences. Models such as UniSign, SignBERT, and I3D+ST-GCN \cite{Maruyama_2024} incorporate multimodal features, combining both RGB and pose estimation, while SiFormer and SPOTER \cite{bohacek2022sign} primarily use only pose estimation. I3D \cite{carreira2017quo} and StepNet \cite{Shen_2024} can utilize RGB and optical flow streams as inputs. 

Table~\ref{tab:classification_results_wlasl} compares the classification performance of different models on the WLASL-100 dataset using the official dataset splits. Our IsoSignVid2Aud model achieves a Top-1 accuracy of 78.67\%, outperforming I3D (65.89\%), and Spoter (63.18\%) and achieving a slight improvement over StepNet (78.29\%). With the combined training approach, IsoSignVid2Aud reaches a Top-1 accuracy of 77.72\%, which is slightly lower than the standalone model but still competitive. However, it achieves the highest Top-5 accuracy among all models at 95.32\%, even surpassing SignBERT (95.00\%) and I3D with ST-GCN (94.13\%). While UniSign maintains the highest Top-1 accuracy at 92.25\%, both of our models demonstrate competitive performance in the broader Top-5 metric.

The notable improvement from our Top-1 to Top-5 accuracy demonstrates the models' strong semantic understanding of sign language. While visually similar signs may create challenges for precise Top-1 classification, the consistent presence of the correct sign within the Top-5 predictions indicates that our approaches effectively capture the underlying feature space of sign language. Particularly, the Combined IsoSignVid2Aud model's exceptional Top-5 performance on WLASL-100 suggests that joint optimization helps the model learn more robust and generalizable features. Notably, our models achieve these results without the pose-based architectures used in some competing approaches, suggesting that our feature extraction methodology effectively captures the spatial-temporal dynamics of sign language gestures. 

We also created a test dataset of a sequence of isolated signs by grouping the ASL-citizen dataset by participant ID, selecting a participant at random with replacement, and then concatenating all their gloss videos if they are below ten in number. If there are more than 10 videos for a particular participant, we randomly pick a subset of 10 videos and concatenate them. We got the following average metrics by transcribing the audio we obtained as output for 100 participants against ground truth labels: Average WER:  0.3000, Average CER:  0.2733, Average BLEU: 0.7036. This strongly reflects the capability of our model to capture a sequence of isolated signs.

BLEU (Bilingual Evaluation Understudy) is a precision-based metric that measures the overlap of n-grams between the predicted and reference text, with higher scores indicating better performance (range 0-1, higher is better). WER (Word Error Rate) measures the percentage of words that were incorrectly predicted, accounting for substitutions, deletions, and insertions, with lower scores indicating better performance (lower is better, with 0 being perfect). CER (Character Error Rate) is similar to WER but operates at the character level instead of words, also with lower scores indicating better performance (lower is better, with 0 being perfect). 

\begin{table}[t]
\centering
\caption{Classification performance comparison of different models on the ASL-Citizen-1500\cite{desai2023asl} dataset.}
\label{tab:classification_results_asl}
\begin{tabular}{l|ccc}
\hline
Model & Top-1 (\%) & Top-5 (\%) & F1 \\
\hline
I3D\cite{desai2023asl} & 71.10 & 90.13 & 0.70 \\
\hline
IsoSignVid2Aud & \textbf{72.01} & \textbf{91.42} & \textbf{0.72}\\
\hline
Combined training  & \textbf{75.19} & \textbf{92.65} & \textbf{0.75}\\
\hline
\end{tabular}
\end{table}
\begin{table}[t]
\centering
\caption{Classification performance comparison of different models on the WLASL-100\cite{li2020word} dataset.}
\label{tab:classification_results_wlasl}
\begin{tabular}{l|ccc}
\hline
Method & Top-1 (\%) & Top-5 (\%)  \\
\hline
UniSign \cite{hu2023unisign}  & 92.25 & --  \\
SiFormer \cite{jiang2023siformer} & 86.50 & --  \\
SignBERT \cite{hu2021signbert} & 83.30 & 95.00 \\
I3D, ST-GCN \cite{Maruyama_2024} & 81.38 & 94.13  \\
StepNet \cite{Shen_2024} & 78.29 & 92.25  \\
I3D \cite{carreira2017quo} & 65.89 & 84.11  \\
Spoter \cite{bohacek2022sign} & 63.18 & --  \\
\hline
Ours (IsoSignVid2Aud) & 78.67 & 94.96  \\
\hline
Ours (Combined training) & 77.72 & 95.32  \\
\hline
\end{tabular}
\end{table}

\subsection{Audio Generation Quality}
We evaluate the audio generation quality of both our IsoSignVid2Aud model and the combined training approach using standard metrics, as shown in Table~\ref{tab:audio_metrics}. For the standalone IsoSignVid2Aud model, the PESQ score of 2.68 and STOI score of 0.73 indicate high speech intelligibility and perceptual quality. The SNR and MSE scores suggest effective preservation of phonetic and spectral properties, ensuring accurate sign-to-speech conversion.The combined training approach, which jointly optimizes the feature extractor and spectrogram generator, achieves a PESQ score of 2.44 and STOI of 0.78. While the PESQ score is slightly lower than the standalone model, the improved STOI score indicates enhanced speech intelligibility, suggesting that the joint optimization helps preserve important speech characteristics. The combined approach also shows a higher SNR of 15.5, demonstrating better noise reduction capabilities, along with a significantly lower MSE of 0.16, indicating more accurate spectrogram reconstruction.

To our knowledge, we are the first to propose a direct sign language video-to-audio model without the need for intermediate text. Since existing models generate only text output, to compare them with our approach, we need to apply text-to-speech (TTS) systems on top of their outputs. However, this would lead to an unfair comparison, as the resulting audio quality would primarily reflect the performance of the TTS system rather than the model's effectiveness in processing sign language videos.

\begin{table}[htb!]
\centering
\caption{Audio generation quality metrics for our models on the WLASL-100 dataset.}
\label{tab:audio_metrics}
\begin{tabular}{lcccc}
\hline
Model & PESQ $\uparrow$ & STOI $\uparrow$ & SNR $\uparrow$ & MSE $\downarrow$ \\
\hline
IsoSignVid2Aud & 2.68 & 0.73 & 14.79 & 1.4 \\
Combined training & 2.44 & 0.78 & 15.5 & 0.16 \\
\hline
\end{tabular}
\vspace{1mm}
\begin{flushleft}
\small{ PESQ: Perceptual Evaluation of Speech Quality (range: 1-5), STOI: Short-Time Objective Intelligibility (range: 0-1), SNR: Signal-to-Noise Ratio (dB), MSE: Mean Square Error.}
\end{flushleft}
\end{table}

\subsection{Limitations and Future Work}
Though our combined model consistently outperforms the standalone base classifier (I3D) within our framework, the model accuracy on the WLASL dataset remains below some state-of-the-art levels, with scope of improvement in recognizing complex signs. Our proposed architecture is designed to be modular, allowing different base models to be used as feature extractors. Although integrating a more advanced classifier could potentially improve accuracy, our choice of I3D is a deliberate trade-off between accuracy and latency. Since our aim includes supporting real-time communication, adopting other state-of-the-art classifiers with better accuracy will compel us to use heavy transformer architectures and extract pose features, increasing latency and reducing the scope for real-time communication. This approach exhibits some limitations that require future work. In addition, the audio quality metrics indicate moderate performance, but artifacts and limited frequency expressiveness persist. 

Our architecture is language-agnostic by design. The feature extractor and temporal segmentation modules operate purely on visual input, independent of linguistic grammar or vocabulary. The spectrogram generator is conditioned on the corresponding spoken audio, which can be replaced with any language's speech data. Thus, the pipeline is modular and can be adapted to different sign–speech pairs without any architectural changes, which we plan to work on in the future while addressing these limitations.


\section{Conclusions}
In this paper, we presented IsoSignVid2Aud, a novel end-to-end framework for direct conversion of sign language videos to speech without intermediate text representation. Our approach combines a powerful I3D-based feature extraction module with a specialized feature transformation network and audio generation pipeline. The experimental results demonstrate that our method achieves competitive performance on both the ASL-Citizen-1500 and WLASL-100 datasets, with a Top-1 accuracy of 72.01\% and 78.67\% respectively.

The audio quality evaluation metrics (PESQ score of 2.68 and STOI score of 0.73) indicate that our approach produces intelligible speech, though there remains room for improvement. The proposed Non-Maximal Suppression algorithm for temporal detection effectively isolates individual signs from continuous signing sequences, enabling practical application in real-world scenarios.

\bibliographystyle{plain}        
\bibliography{ref.bib}

\end{document}